\documentclass[letterpaper, 10 pt, conference]{ieeeconf}  

\IEEEoverridecommandlockouts                              

\usepackage{tabularray}
\usepackage{soul}
\usepackage{cite}
\usepackage{svg}
\usepackage{amsmath,amssymb,amsfonts}
\usepackage{algorithmic}
\usepackage{graphicx}
\usepackage{textcomp}
\usepackage{xcolor}
\usepackage{dsfont}
\usepackage{subfigure}
\usepackage{booktabs}
\usepackage{bbm}
\usepackage{url}
\usepackage{svg}
\usepackage{multirow}
\usepackage{multicol}
\usepackage{rotating}
\usepackage{color}
\usepackage{bigstrut}
\usepackage{array}
\usepackage{tabularx}
\usepackage{amssymb}
\usepackage{pifont}

\usepackage[normalem]{ulem}
\useunder{\uline}{\ul}{}
\usepackage{marvosym}
\usepackage{color, xcolor} 
\usepackage{placeins}
\usepackage{afterpage}

\makeatletter
\let\NAT@parse\undefined
\makeatother
\usepackage{hyperref}
\usepackage[nameinlink,noabbrev]{cleveref}

\title{\LARGE \bf
Bench4Merge: A Comprehensive Benchmark for Merging in Realistic Dense Traffic with Micro-Interactive Vehicles
}

\author{Zhengming Wang$^{1,2}$, Junli Wang$^{3}$, Pengfei Li$^{1}$,  Zhaohan Li$^{1}$, Chunyang Liu$^{4}$, Bo Zhang$^{4}$\textsuperscript{\Letter}, Peng Li$^{1}$\textsuperscript{\Letter}, Yilun Chen$^{1}$ 
\thanks{$^{1}$ Institute for AI Industry Research (AIR), Tsinghua University, China
        \{peng-li,Chenyilun\}@air.tsinghua.edu.cn,li-pf22@mails.tsinghua.edu.cn}%
\thanks{$^{2}$College of Energy Engineering, Zhejiang University, China,
        wzm5853@zju.edu.cn.}%
\thanks{$^{3}$ Institute of Automation, Chinese Academy of Sciences, China.}%
\thanks{$^{4}$ Didi Chuxing, China.}%
}

\begin{document}

\maketitle
\thispagestyle{empty}
\pagestyle{empty}

\begin{abstract}

While the capabilities of autonomous driving have advanced rapidly, merging into dense traffic remains a significant challenge, many motion planning methods for this scenario have been proposed but it is hard to evaluate them. Most existing closed-loop simulators rely on rule-based controls for other vehicles, which results in a lack of diversity and randomness, thus failing to accurately assess the motion planning capabilities in highly interactive scenarios. Moreover, traditional evaluation metrics are insufficient for comprehensively evaluating the performance of merging in dense traffic. In response, we proposed a closed-loop evaluation benchmark for assessing motion planning capabilities in merging scenarios. Our approach involves other vehicles trained in large scale datasets with micro-behavioral characteristics that significantly enhance the complexity and diversity. Additionally, we have restructured the evaluation mechanism by leveraging Large Language Models (LLMs) to assess each autonomous vehicle merging onto the main lane. Extensive experiments and test-vehicle deployment have demonstrated the progressiveness of this benchmark. Through this benchmark, we have obtained an evaluation of existing methods and identified common issues. The simulation environment and evaluation process can be accessed at \url{https://github.com/WZM5853/Bench4Merge}.

\end{abstract}


\section{Introduction}

Merging into dense traffic is a typical scenario of highly interactive driving, where autonomous vehicles face significant challenges \cite{knaup2024active}. The difficulty often arises from the inability to effectively interact with surrounding vehicles to create sufficient space for merging, leading to stagnation. Recently, there has been growing research on motion planning in such scenarios \cite{hou2024merging,brito2022learning,bouton2019cooperation}, with the hope to enhance the traffic-handling capability in dense merging scenarios. However, a comprehensive evaluation of these methods in simulation remains an unresolved issue. This is primarily due to the interactivity of the merging process, the lack of realistic micro-level interaction behaviors in surrounding vehicles limits further analysis, training, and ultimately, the optimization of these methods.

\begin{figure}[tpb]
\centerline{\includegraphics[width=0.5\textwidth]{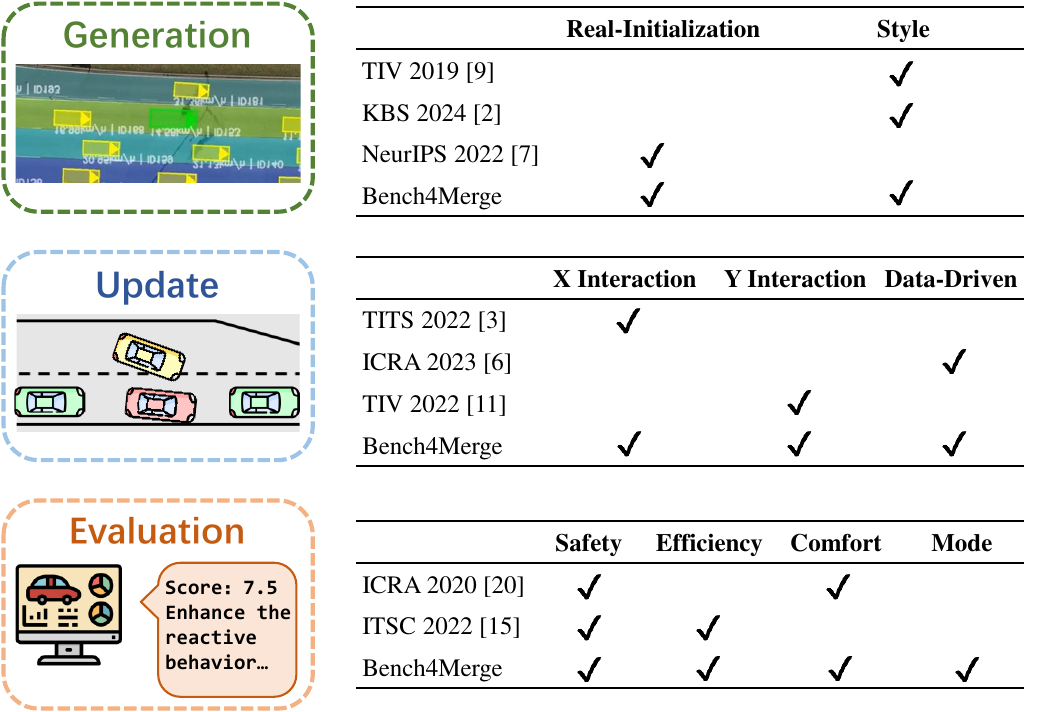}}
\caption{A closed-loop benchmark includes three components, respectively, Generation, Update, Evaluation. “Real-Initialization” indicates that the initial scenario is extracted from real data; “Style” represents different vehicle styles; “X Interaction” and “Y Interaction” represent the vehicle's lateral and longitudinal displacements, respectively; “Safe.”, “Effi.”, and “Comf.” denote the three evaluation metrics of safety, efficiency, and comfort; and “Mode” signifies the vehicle's mode, including hurry, medium and relax.}
\label{fig:1}
\end{figure}

        
        

Typical closed-loop benchmarks for autonomous driving consists of three components: \textit{initial scenario generation}, \textit{scenario iterative updating}, and \textit{evaluation metrics}, Fig.~\ref{fig:1} shows the comparison. In existing studies, generating initial scenarios can be categorized into rule-based arrangements and neural network-based generation. Rule-based methods typically adopt predefined definitions of dense traffic from previous research \cite{hou2024merging, ni2016vehicle}. However, those approaches struggle to reflect the realism of the environment \cite{xu2023bits}. Some studies have introduced traffic generation from real-world data, but fail to distinct different styles \cite{gulino2023waymax}. On the other hand, neural network-based methods for generating initial states usually involve updating the environment over a period of time \cite{suo2021trafficsim}, during which the interaction between vehicles is not well-represented, makes those unsuitable for closed-loop simulations in dense merging scenarios \cite{nishi2019merging}. Therefore, a realistic and interactive environment is essential.

The primary aspect of iterative scenario updates is the state updating of surrounding vehicles. In closed-loop simulation environments, vehicle movements are typically determined using rule-based methods \cite{dauner2023parting}. Efforts have been made to improve these rules to achieve more diverse behaviors \cite{brito2022learning}. Researchers have realized the importance of lateral displacements (Y Interaction) in micro-level interactions \cite{hasuo2022goal}, but haven't introducted it in merging scenarios, as shown in Fig.~\ref{fig:1}, which is a key behavior in dense traffic \cite{chandra2023meteor}. Data-driven approaches have increasingly been used for surrounding vehicles, but as mentioned earlier, these generated behaviors are often in the form of pre-determined trajectories, making them unsuitable for real-time interaction \cite{xu2023bits}. Iterative update planning methods have gained attention as they address the limitations regarding interactions \cite{wang2023multiverse}. Nevertheless, these methods still struggle to capture the diversity of behaviors, as neural networks tend to produce averaged strategies. Therefore, the simulation of dense merging scenario requires surrounding vehicles that exhibit microscopic interaction characteristics.

The design of evaluation metrics has a profound impact on the final assessment of algorithms. Traditionally, metrics have been discretely categorized into three aspects: safety, efficiency, and comfort \cite{kuutti2021weakly,liu2022autonomous,liu2022zju}. Existing studies select some of those aspects, as shown in Fig.~\ref{fig:1}, and fail to consider the drive mode, such as the vehicle is in a hurry or relax. For instance, sharp acceleration rate can improve merging efficiency but significantly compromise safety and comfort, and drive mode also affect the behaviors. This reliance on a single metric leads to a one-sided evaluation of algorithms, failing to capture their overall performance. Therefore, a comprehensive evaluation framework is required to analyze the performance of different methods.

In this work, we propose a new benchmark named Bench4Merge for evaluating merging in realistic dense traffic, leveraging large-scale data to extract realistic initial states and classify the initial scenarios to select more targeted environments. We developed a deep neural network architecture to capture the micro-level interaction behaviors from the data. The vehicles update their planning and state based on the frequency of the environment. By incorporating vectorized labels, we achieve diverse interaction styles, addressing the previous limitations of interaction and behavioral diversity in surrounding vehicles. To enable a comprehensive evaluation, we reconstruct the evaluation mechanism, utilizing LLMs to score and assess the entire merging process. This approach overcomes the limitations of previous evaluations that relied on one-sided metrics.

To summarize, our contributions are as follows:

\begin{itemize}
\item[$\bullet$] We highlight key challenges in evaluating motion planning methods for dense merging scenarios.
\item[$\bullet$] We propose a new benchmark for merging, named Bench4Merge, which incorporates more realistic merging scenarios, richer microscopic interactive behaviors, and more comprehensive evaluation mechanism.
\item[$\bullet$] We have implemented various motion planning algorithms for merging scenarios, uncovering issues that were previously overlooked in evaluations. The complete project has been open-sourced.
\end{itemize}

\section{Related Works}
\textbf{Closed-Loop Simulation for Merging Scenarios.} 
The IDM-based approach has been validated to effectively simulate the longitudinal behavior of vehicles \cite{burger2022interaction}. Initially, researchers used improved IDM-based (Intelligent Driver Model) methods to control the surrounding vehicles, capturing the diversity of behaviors \cite{bouton2019cooperation} and adding a prediction module \cite{brito2022learning}, allowing surrounding vehicles to exhibit reactive capabilities. Some also utilized the real data to cluster different types of vehicles \cite{hou2024merging}. However, this approach is limited in its inability to simulate the microscopic lateral behaviors. Additionally, they conducted a classification of vehicle styles but did not perform a classification of the associated scenarios. To capture realistic microscopic behaviors, real-world dataset has been used to extract vehicle behavior characteristics \cite{zhang2023trafficbots}, and diffusion models has been applied to generate diverse scenarios \cite{zhong2023guided}, yielding promising results in traffic scenario generation \cite{suo2021trafficsim}. However, since the generated trajectories lack interactivity, they cannot be applied in closed-loop simulations \cite{jia2024bench2drive}. Iterative generation methods, which allow other vehicles planning to update with the frequency of the environment \cite{wang2023multiverse}. By encoding historical trajectories, maps, and other information into neural networks, real-time planning and execution for environment vehicles can be achieved \cite{zhou2024behaviorgpt}. We extract the initial scenarios from a well-classified set, building on the capability for closed-loop updates to reflect interactivity, meanwhile, we employed additional data labeling to enable the vehicle to exhibit different interaction styles.

\textbf{Evaluation Metrics for Merging in Dense Traffic.} 
The evaluation metrics for motion planning in dense merging scenarios are generally categorized into safety, efficiency, and comfort. Safety can be further divided into explicit and implicit safety. Collision rates represent explicit safety, while metrics TTC (Time to Collision) and TH (Time Headway) indicate implicit safety \cite{bae2022lane,lee2023robust}. For traffic efficiency, metrics such as average travel time, timeout rates and average speed\cite{ding2021epsilon,li2023marc,gu2023exploring} are used. There has been a growing interest in the microscopic aspects of merging maneuvers, including the distribution of merging zones and the distance traveled before merging \cite{hao2020research,daamen2010empirical}. Additionally, some researchers have analyzed the fluctuations in overall traffic flow caused by merging behavior to assess the impact of the merging vehicle \cite{wang2023faster}. For comfort, initial assessments often relying on defining acceleration ranges \cite{zhang2020efficient} or calculating Jerk to measure fluctuations in vehicle acceleration \cite{liu2022zju}. Most of the existing research primarily considers only a single metric. Some studies have considered multiple metrics and assigned weights to each \cite{dauner2024navsim}, yet the specific values of these weights remain highly arbitrary. Our evaluation are more comprehensive and better aligned with human evaluations.


\section{Bench4Merge}

\begin{figure*}[t]
\centerline{\includegraphics[width=\textwidth]{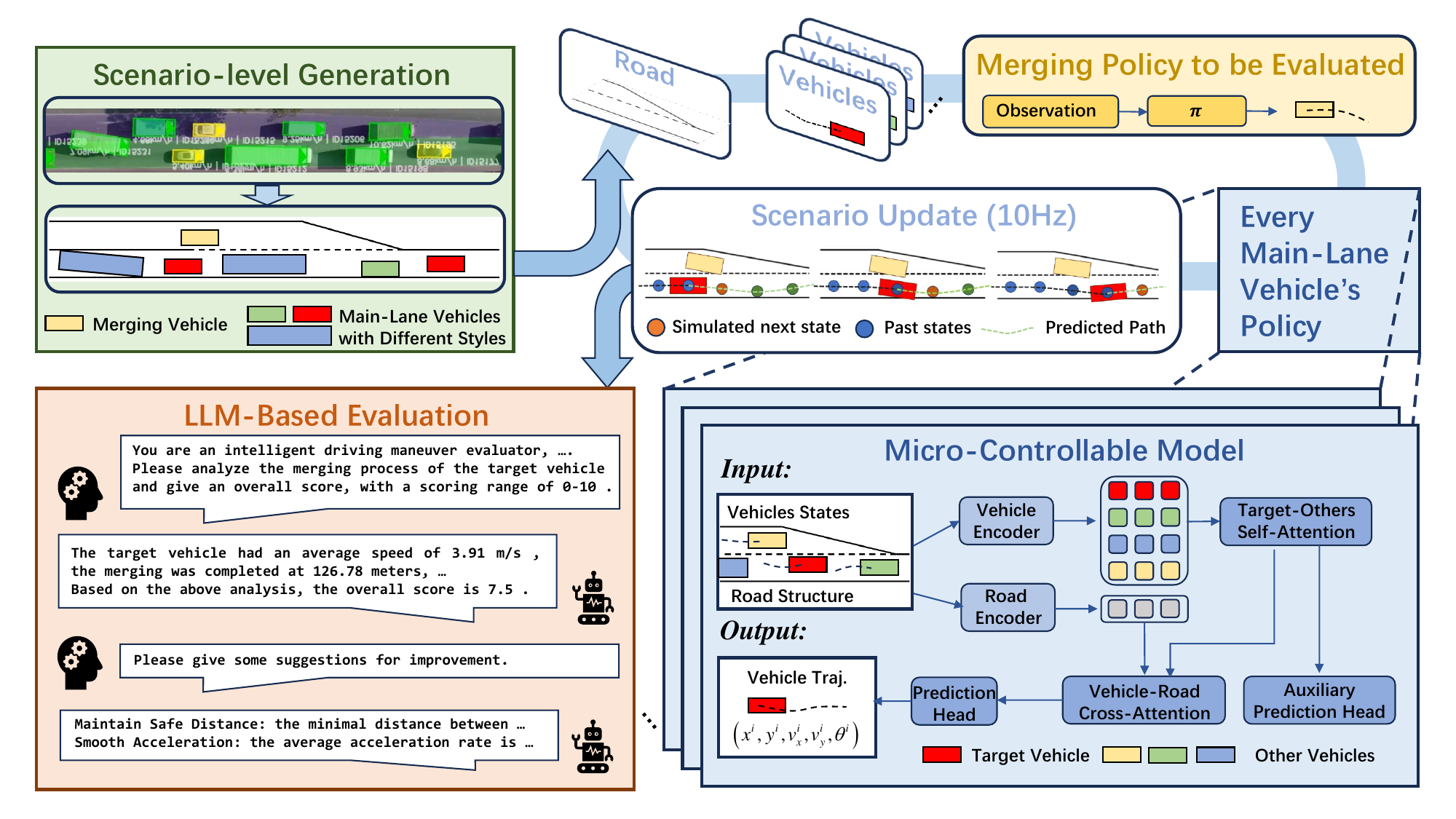}}
\caption{\textbf{Overview of our architecture.} Bench4Merge consists of three main parts: Scenario-level Generation, Micro-Controllable Model for main-road vehicles, and LLM-Based Evaluation. In this context, Merging Policy refers to the planning method of the merging vehicle being evaluated.}
\label{fig:2}
\vspace{-0.5cm}
\end{figure*}

\subsection{Overview}

Bench4Merge consists of three main components: \textit{Scenario-level Generation}, \textit{Micro-Controllable Model} for Main-Lane Vehicles, and \textit{LLM-Based Evaluation}, as shown in Fig.~\ref{fig:2}. Firstly, unlike the method of randomly arranging the initial scene based on predefined rules, our scenarios are derived entirely from classified real-world data. Secondly, the motion policy of surrounding vehicles are trained on large-scale dense merging traffic datasets, effectively capturing micro-level interactive characteristics. We incorporate feature tags to reflect different vehicle personalities, enabling each vehicle in the environment to individually observe states, plan trajectories, and dynamically update them iteratively to achieve real-time interactions. Finally, the final evaluation module is based on an LLM. Our approach leverages the LLM to score the merging process of the tested vehicle and provide improvement suggestions. This method addresses the limitations of previous discrete metrics and allows for dynamic adjustments based on the driving mode of a vehicle.

\subsection{Initial Scenarios}

\begin{figure}[tpb]
\centerline{\includegraphics[width=0.47\textwidth]{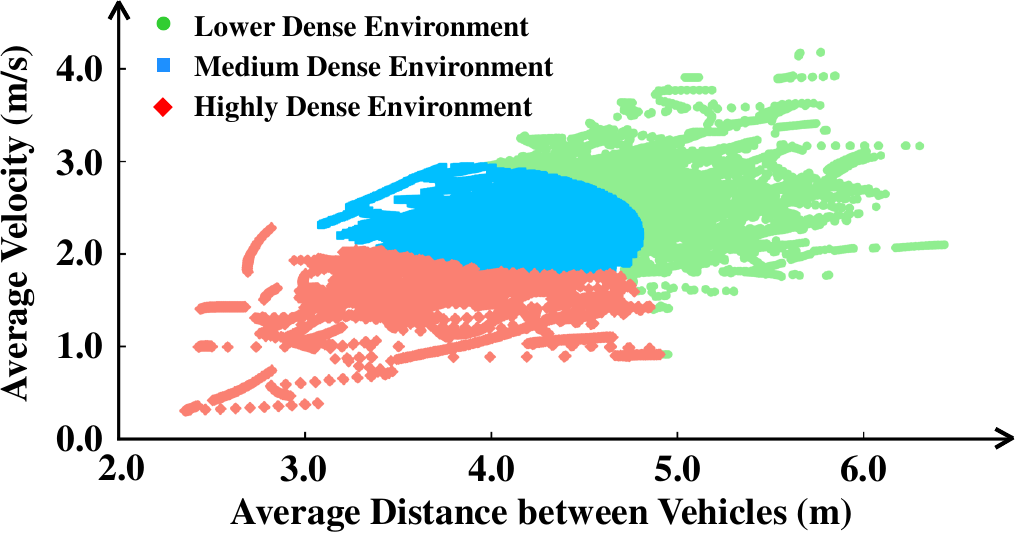}}
\caption{The average speed and average distance distribution of other vehicles in each init environment, we divide them into three categories.}
\label{fig:3}
\end{figure}

To achieve realistic initial environment generation, we extracted initial scenarios from the real world \cite{zhang2023ad4che} and classified these scenarios to provide Bench4Merge with a more diverse set of scenes for comprehensive evaluation of different methods. This approach overcomes the limitations where vehicles were merely arranged along the centerline according to predefined rules \cite{bouton2019cooperation,brito2022learning}, lacking realism and diversity in initial environment generation.

Ultimately, we extracted over 50,000 initial scenarios from the DJI Dense Traffic Dataset \cite{zhang2023ad4che}. As shown in Fig.~\ref{fig:3}, it is evident that the average speed and average distance in each scenario exhibit a linear relationship: the lower the average spacing, the slower the average speed. We classified the scenarios using the Gaussian Mixture Model (GMM) \cite{lao2012gaussian}, with the average speed and average spacing as classification features, dividing the scenarios into three categories: \textit{highly dense environments}, \textit{medium dense environments}, and \textit{lower dense environments}. Table.~\ref{tab.5} in Section \uppercase\expandafter{\romannumeral4} demonstrates the significance of classifying these environments.

\subsection{Micro-Controllable Vehicles Model}
\begin{figure}[tpb]
\centerline{\includegraphics[width=0.5\textwidth]{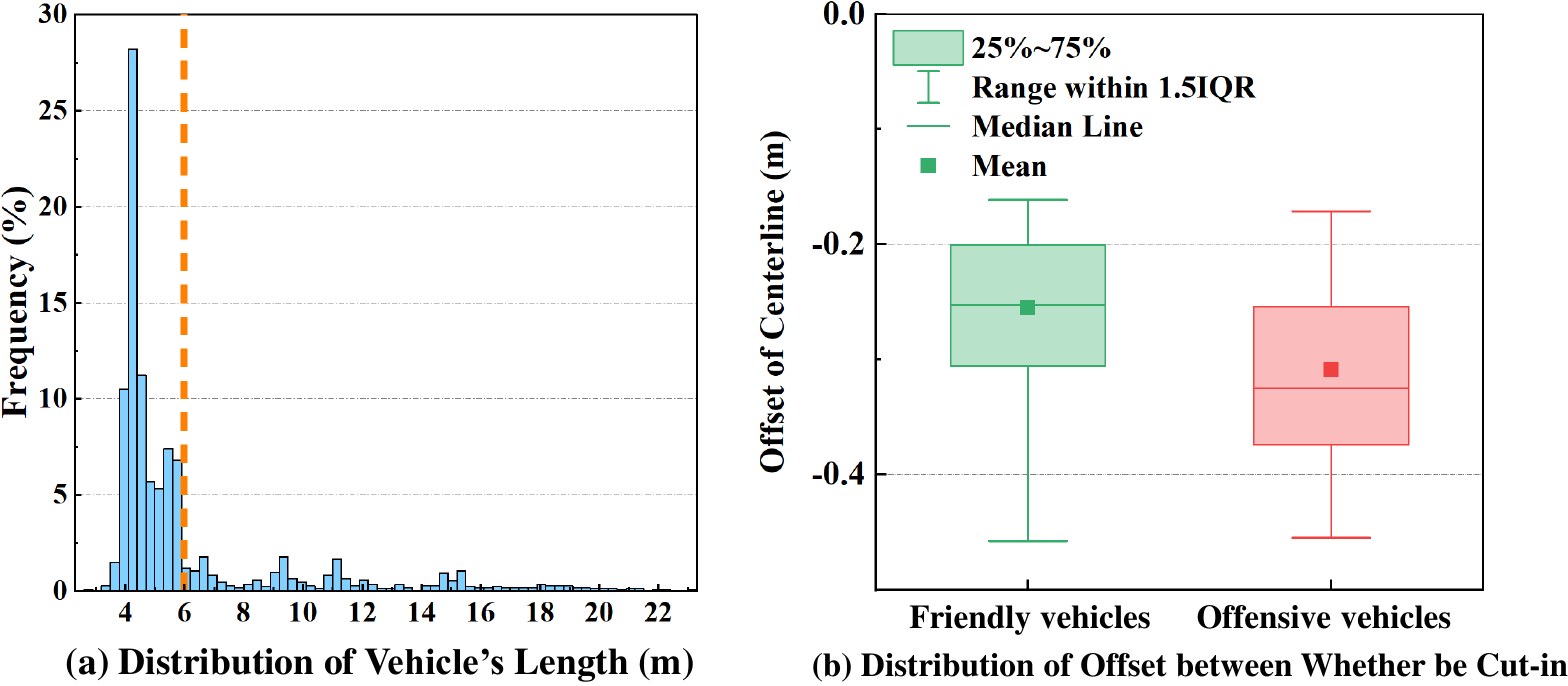}}
\caption{Analysis of the datasets}
\label{fig:4}
\end{figure}

\textbf{Training Data Construction.}
We selected the DJI Dense Traffic Dataset as the primary data source \cite{zhang2023ad4che}. To enhance data diversity and improve the generalization capability, we also extracted a significant amount of related data from merging scenarios within the nuPlan \cite{krajewski2018highd}, ExiD \cite{caesar2021nuplan} and our own collected datasets, the specific screening criteria can be found in the appendix. To better represent a variety of behavioral types among those vehicles, we conducted an analysis based on DJI dataset, for it has the most comprehensive vehicle classification, we classified the vehicles into three categories: \textit{long vehicles}, \textit{offensive vehicles}, and \textit{friendly vehicles}, as these three types exhibit significant differences in interaction behaviors. 
Fig.~\ref{fig:4}(a) shows the distribution of vehicle lengths, according to Chinese traffic regulations, the dividing line between long and short vehicles is defined as 6$m$, with average length of 4.7$m$ and 11.5$m$. Both offensive and friendly vehicles fall into the category of short vehicles.

We define vehicles that have not been cut in line throughout the entire process as offensive vehicles, while others are classified as friendly vehicles. As shown in Fig.~\ref{fig:4}(b), offensive vehicles exhibit a greater degree of lateral deviation from the centerline, whereas friendly vehicles demonstrate smaller lateral shifts. This further reinforces the significance of microscopic lateral interactions exhibited by the main road vehicles. Table.~\ref{tab.4} in Chapter \uppercase\expandafter{\romannumeral4} summarizes the specific data, long vehicles exhibit distinct characteristics compared to the other two types.

\begin{figure}[tpb]
\centerline{\includegraphics[width=0.5\textwidth]{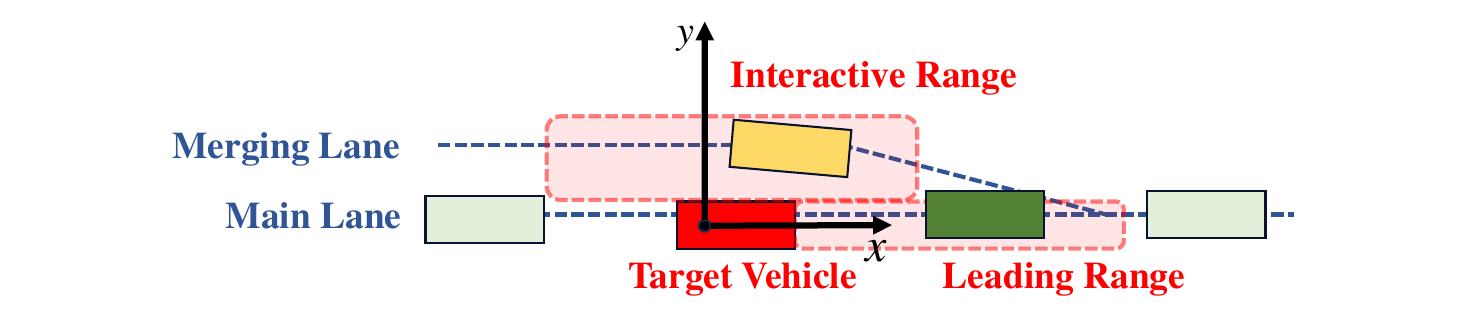}}
\caption{Every sample includes the state of the target vehicle and other vehicles, as well as map information. The other vehicles are all those fall within the leading and interaction range of the target vehicle.}
\label{fig:5}
\end{figure}
Finally, we constructed each training sample according to Equation.~\ref{equ:1}. Each sample includes the state of the target vehicle and other vehicles, as well as map information, as shown in Fig.~\ref{fig:5}. The “other vehicles" category includes all vehicles that fall within the target vehicle's leading range and interactive range.
\begin{equation}
\begin{split}
samp_{i}= \left\{ 
\begin{array}{ll}
veh_{target} & = \left\{ x,y,\theta, v_{x},v_{y},a_{x},a_{y},...,label \right\} \\[0.2cm] 
veh_{other}^{i} & = \left\{ x,y,\theta, v_{x},v_{y} \right\} \\[0.1cm]
road & = \left\{ x_{main}^{j},y_{main}^{j},x_{merge}^{j},y_{merge}^{j} \right\} \\ 
\end{array} 
\right.
\end{split}
\label{equ:1}
\end{equation}
where the coordinate system is centered at the rear axle of the target vehicle, $\theta$ represents the heading angle, $v$ denotes the velocity, $a$ is the acceleration, and $d$ indicates the distance to the vehicle ahead. The $label$ corresponds to the vehicle's style labels. $x_{main}^{j}$, $y_{main}^{j}$ are the waypoints of main lane and $x_{merge}^{j}$, $y_{merge}^{j}$ are the waypints of merging lane. Our environment updates at a frequency of 10 Hz, with each data sample comprising 50 frames. The first 10 frames serve as input, while the subsequent 40 frames are used as ground truth. Our vehicle classification method can be adapted to more styles and more details can be found in appendix.

\textbf{Training Setting.}
We design an imitation-learning-based model, to simulate the driving behaviours of the main-lane vehicles, which is shown in the bottom right part of Fig.~\ref{fig:2}. 
Aiming at effectively capturing micro-level interactive characteristics, the model takes all the vehicles states $V=\{V^i\}_{i=1}^{N_v}$, including selected vehicle $V_s$ and main-lane vehicle $V_m$, and road lanes $R=\{R_i\}_{i=1}^{N_r}$ as input, planning the future trajectory for each main-lane vehicle based on attention mechanisms. Here, ${{N}_{v}}$ and ${{N}_{r}}$ represent the number of vehicle and road polylines.

Specifically, we represent each vehicle state as $V^i \in \mathbb{R}^{T_{h}\times D_{v}}$ and road polyline as $R^i \in \mathbb{R}^{N_r \times D_{r}}$, where $T_{his}$ and $N_r$ correspond to the history frame count and polyline point count, respectively. $D_{v}$ and $D_{r}$ are the dimensions of the input features, where $D_{v}$ consists of the position, velocity, acceleration and the personality label of each vehicle at each timestamp, while $D_{r}$ contains the coordinates of each point.
We first normalize these features to the local coordinate system of a target main-lane vehicle $V_j$. Then, the vehicle features and road features are flattened and projected to the same dimension $D$ with two specific linear networks for future attention operation:
\begin{equation}
\begin{split}
F_v = \mathrm{Linear}_v(\{V^i\}_{i=1}^{N_v}), F_r = \mathrm{Linear}_r(\{R_i\}_{i=1}^{N_r}).
\end{split}
\end{equation}

Then, the projected features $F_v\in {{\mathbb{R}}^{{{N}_{v}}\times D}}$ and $F_r\in {{\mathbb{R}}^{{{N}_{r}}\times D}}$ are fed into two attention-based modules. A module built with self-attention is first leveraged for $F_v$, in which all the vehicle features interact with each other, such that the micro-level interactive characteristics of the vehicles can be captured. In the second module, the updated vehicle features serve as queries and the road features serve as keys and values, together fed into the cross-attention layers. After that, the vehicle features obtain the information of both other vehicles and road structure.
\begin{equation}
\begin{split}
F_v^{\prime} = \mathrm{CrossAttn}(\mathrm{SelfAttn}(F_v, F_v), F_r).
\end{split}
\end{equation}

Finally, another linear network is used to predict the future trajectory $P_{pred}^{j}\in\mathbb{R}^{T_{fut}\times D_{v}}$ of the target vehicle:
\begin{equation}
\begin{split}
P_{pred}^{j} = \mathrm{Linear}_{out}(F_{v_j}^{\prime}).
\end{split}
\end{equation}
Note that we only take the prediction result of the target vehicle for simulation to keep the coordinates consistent. After processing each main-lane vehicle, the whole scenario can be updated.

During training, a driving scene is split into several data samples, each containing one target vehicle. For each data sample, we employ the Mean Squared Error (MSE) loss to supervise the predicted future positions, speeds, and heading angles of the target vehicle:
\begin{equation}
\begin{split}
\mathcal{L}_{tar}=\frac{1}{T_{fut}}{\sum\limits_{t=0}^{T_{fut}-1}{\lambda \left( t \right)\left\| P_{pred}^{j}(t)-P_{gt}^{j}(t) \right\|}^{2}}.
\end{split}
\end{equation}
where $\lambda \left( t \right)$ represents an exponentially decaying weighting mechanism, which allows the prediction errors at different time steps to be weighted according to their significance:
\begin{equation}
\begin{split}
\lambda \left( t \right)={{e}^{\frac{t-{{T}_{fut}}}{t}}}+1.
\end{split}
\end{equation}
Along with the loss for the target vehicle, an auxiliary prediction module is used to output trajectories of all vehicles and losses for each vehicle are computed for efficient training. Thus, the final loss is composed as follows:
\begin{equation}
\begin{split}
\mathcal{L}_{total}={{\gamma }_{1}}\mathcal{L}_{tar} + {{\gamma }_{2}}\mathcal{L}_{aux}.
\end{split}
\end{equation}

\subsection{LLM-Based Comprehensive Evaluation}

\begin{figure}[tpb]
\centerline{\includegraphics[width=0.5\textwidth]{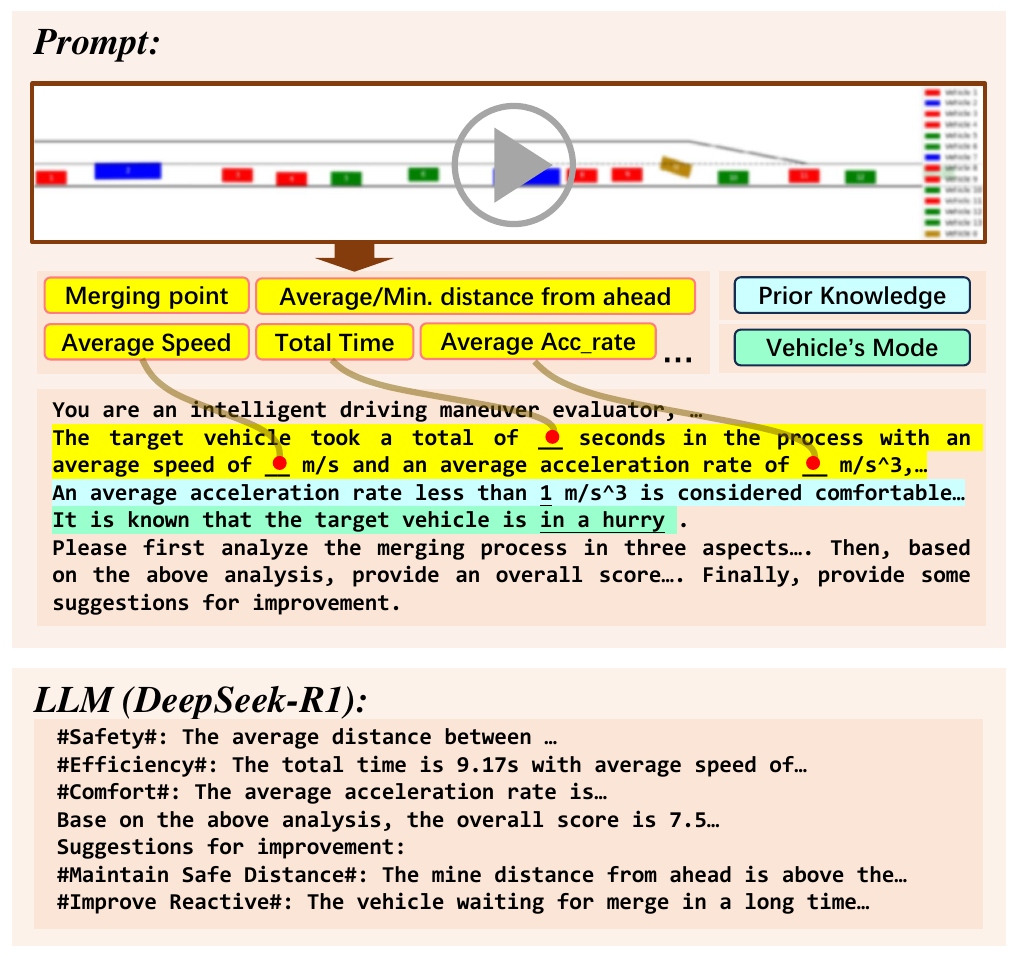}}
\caption{The LLM-based evaluator}
\label{fig:6}
\end{figure}

As shown in Fig.~\ref{fig:6}, we first compute data for each sample, including the average speed, the average/maximum changing rate of acceleration, total time taken, the location of the merging point, and the average/minimum distances to other vehicles, and we also input the main-lane traffic info. Additionally, we input key prior knowledge, the comfort acceleration range, effectively speed range \cite{li2023marc} and the safe time gap with other vehicles, which provide the LLM with necessary references. Finally, we also input the mode of the tested vehicle, for instance, whether the vehicle is currently in a hurry or relax. This is a significant advantage of using an LLM, as it eliminates the need for designing complex rules to evaluate vehicles under different modes \cite{sha2023languagempc}, a challenge in previous approaches.
 
In the design of the prompt, we first instruct the model to analyze the scenario from three perspectives: safety, comfort, and efficiency. Then, it is asked to provide an overall score, with explicit instructions not to assign individual scores for each aspect but to focus on a comprehensive score. Finally, the model is prompted to suggest potential improvements for the method. This approach is intended to intuitively identify the shortcomings of the method and to uncover common issues across various methods, thereby guiding further optimization of the planning methods. We have deployed this system on the real testing-platform and provided a demonstration in the appendix.


\section{Experiments}

\subsection{Effectiveness of LLM-Based Evaluation}
We introduce an LLM-based evaluator, specifically, DeepSeek-R1 671B \cite{Deepseekr1}. To demonstrate the effectiveness of LLM as an evaluation mechanism, we introduced a comparison with human experts like the previous work \cite{chen2024driving}. We conducted 100 experiments, saving both video and data, and presented them to human experts in the same format as they were input to the LLM. We invited 10 experts, dividing them into five pairs, with each pair of experts responsible for scoring the same set of 20 samples, as shown in Table.~\ref{tab.2}. We then compared the scores from the experts with those from the LLM. The Pearson correlation coefficient and MSE were used to assess the correlation and differences between the human scores and the LLM scores, with correlation coefficients exceeding 0.8, which can be considered a strong correlation \cite{asuero2006correlation}. The differences in scores were all within a range of 1 to 2 points.

\begin{table}
\centering
\caption{The Correlation ($\rho$) and MSE Between LLM Scores and Human Expert Scores.}
\resizebox{0.48\textwidth}{!}{
\begin{tblr}{
cells = {c},
cell{1}{2} = {c=2}{},
cell{2}{5} = {c=2}{},
hline{1,3,8} = {-}{},
hline{2} = {2-3}{},
}
        & $\rho$ (Every expert vs LLM) &           & $\rho$               & $\rho$                & MSE    \\
        & Expert\_1                  & Expert\_2 & (Between experts) & ($\text{Expert}_{avg}$ vs LLM)  &        \\
Set1 & 0.9157                     & 0.9382    & 0.9142          & 0.9474            & 1.8305 \\
Set2 & 0.8669                     & 0.8741    & 0.8166          & 0.8788            & 2.2415  \\
Set3 & 0.8286                     & 0.6721    & 0.8422          & 0.8145            & 3.2770  \\
Set4 & 0.9759                     & 0.9868    & 0.9809          & 0.9900            & 0.4108  \\
Set5 & 0.9580                     & 0.9647    & 0.9467          & 0.9653            & 1.0525  
\end{tblr}}
\label{tab.2}
\end{table}

We further demonstrated the effectiveness of the LLM by instructing it to analyze only one specific aspect. We manually analyzed the results, as shown in Table.~\ref{tab.3}, when tasked solely with efficiency, and when the data was adjusted to reflect a more efficient process, the LLM assigned significantly higher scores for the higher speed. Additionally, when the driving style was altered to a “relax" mode while keeping other data constant, the model's assessment of comfort decreased, this shift occurred because the model prioritized comfort over efficiency in this mode, thereby highlighting the comprehensiveness of our evaluation mechanism.

\begin{table}
\centering
\caption{Analysis of Scores Obtained After Modifying the Sample Data and Instructing the LLM to Focus on a Specific Aspect.}
\resizebox{0.48\textwidth}{!}{
\begin{tblr}{
        cells = {c},
        cell{1}{1} = {c=2}{},
        cell{2}{1} = {c=2}{},
        cell{2}{4} = {fg=red},
        cell{3}{1} = {c=2}{},
        cell{3}{4} = {fg=red},
        cell{4}{1} = {c=2}{},
        cell{4}{4} = {fg=red},
        cell{5}{1} = {c=2}{},
        cell{6}{1} = {c=2}{},
        cell{7}{1} = {c=2}{},
        cell{8}{1} = {c=2}{},
        cell{9}{1} = {c=2}{},
        cell{9}{5} = {fg=red},
        cell{10}{1} = {r=3}{},
        hline{1-2,10,13} = {-}{},
        }
        
                &                     & Data1   & Data2         & Data3          \\
        Total time ($s$)   &                     & 8.83   & \textbf{7.00}    & 8.83          \\
        Average speed ($m/s$)   &                     & 4.53    & \textbf{5.71} & 4.53           \\
        Merging point ($m$)     &                     & 124.87  & \textbf{110.00}  & 124.87         \\
        Average jerk ($m/{{s}^{3}}$)   &                     & 0.01 & 0.01       & 0.01        \\
        Max jerk ($m/{{s}^{3}}$)      &                     & 2.95    & 2.95          & 2.95           \\
        …             &                     & …       & …             & …              \\
        Others average speed ($m/s$) &                     & 4.41    & 4.41          & 4.41           \\
        Drive mode   &                     & hurry   & hurry         & \textbf{relax} \\
        LLM Score     & Overall             & 6.7     & 7.2           & 7.8            \\
                        & Only for Efficiency & 6.5     & 9.3           & 6.8            \\
                        & Only for Comfort    & 7.2     & 7.5           & 7.8            
\end{tblr}}
\label{tab.3}
\end{table}

\subsection{Effectiveness of Micro-Controllable Vehicles Model}
Now we demonstrate that the vehicles are capable of reflecting the micro-level characteristics. To this end, we test the environment 100 times, during which we analyzed the differences in key metrics among three vehicle styles we configured. As shown in Table.~\ref{tab.4}, The average distance between friendly and offensive vehicles in the environment is 5.81m and 3.82m, respectively, which closely aligns with 5.35m and 4.07m from the dataset. This demonstrates that our model effectively captures the micro behaviors.

\begin{table}
\centering
\caption{Comparison of the characteristics of different vehicles in the dataset with those in our simulation environment, Offen.,Fri.,Long represent different types we defined before.}
\resizebox{0.48\textwidth}{!}{
\begin{tblr}{
        cells = {c},
        cell{1}{2} = {c=2}{},
        cell{1}{4} = {c=2}{},
        cell{1}{6} = {c=2}{},
        hline{1,3,10} = {-}{},
        hline{2} = {2-7}{},
        vline{4} = {1-9}{solid},
        vline{6} = {1-9}{solid},
}
~                 & Offensive &          & Friendly      &   &  Long        &        \\
~                 & Real    & Ours & Real  & Ours       & Real & Ours   \\
Avg.dhw (m)      & \leavevmode\hphantom{0}4.07   & \leavevmode\hphantom{0}3.82        & \leavevmode\hphantom{0}5.35  & \leavevmode\hphantom{0}5.81   & \leavevmode\hphantom{0}6.49         & \leavevmode\hphantom{0}7.73   \\

Avg.speed (m/s)  &  \leavevmode\hphantom{0}2.96   & \leavevmode\hphantom{0}2.40         & \leavevmode\hphantom{0}2.95   & \leavevmode\hphantom{0}2.15  & \leavevmode\hphantom{0}2.03         & \leavevmode\hphantom{0}1.62   \\

Avg.acc\_x (m/s) & \leavevmode\hphantom{0}0.17   & -0.11        & \leavevmode\hphantom{0}0.19   & -0.08  & \leavevmode\hphantom{0}0.15       & -0.26  \\

Avg.acc\_x\_std & \leavevmode\hphantom{0}0.31    & \leavevmode\hphantom{0}1.91       & \leavevmode\hphantom{0}0.41   & \leavevmode\hphantom{0}1.56  & \leavevmode\hphantom{0}0.09        & \leavevmode\hphantom{0}1.43   \\

Avg.acc\_y (m/s) & -0.07  & -0.07       & \leavevmode\hphantom{0}0.08  & \leavevmode\hphantom{0}0.03   & \leavevmode\hphantom{0}0.04        & \leavevmode\hphantom{0}0.01 \\

Avg.acc\_y\_std & \leavevmode\hphantom{0}0.12   & \leavevmode\hphantom{0}0.09        & \leavevmode\hphantom{0}0.05     & \leavevmode\hphantom{0}0.06 & \leavevmode\hphantom{0}0.04       & \leavevmode\hphantom{0}0.06  \\

Avg.offset (m)   & -0.29   & -0.23       & -0.13  & -0.14   &\leavevmode\hphantom{0}0.04       & \leavevmode\hphantom{0}0.02  
\end{tblr}}
\label{tab.4}
\end{table}

We implemented a RL-based method \cite{chen2023deep} within our environment, the average scores and success rates are lower compared to those in the previous environment as shown in Table.~\ref{tab.5}. Indicating that previously rule-defined main road vehicles lack the micro-level behaviors of real-world traffic participants, resulting in overly simplistic scenarios. Notably, the vehicles exhibited varying success rates across the three different initial scenarios in our environment, underscoring the significance of our initial classification (Section \uppercase\expandafter{\romannumeral3}-B).

\begin{table}
\centering
\caption{Comparison of Success Rates and Scores Across Different Environments Using the Same Merging Planning Method.}
\resizebox{0.48\textwidth}{!}{
\begin{tblr}{
cells = {c},
cell{1}{1} = {c=2}{},
cell{2}{1} = {c=2}{},
cell{3}{1} = {r=3}{},
hline{1-2,6} = {-}{},
}
~                       &                & Success
Rate & Average
Score \\
IDM-based
Environment &                & 85\%           & 5.96            \\
Bench4Merge             & Highly
Dense & 55\%           & 4.99            \\
                        & Medium
Dense & 35\%           & 3.48            \\
                        & Lower
Dense  & 50\%           & 4.42            
\end{tblr}}
\label{tab.5}
\end{table}

\subsection{Evaluation of Representative Methods}
We introduced five representative methods and evaluated them in Bench4Merge. These methods fall into two main categories: optimization-based and neural network-based approaches. For each method, we conducted 100 tests in high-density environments and recorded both the average score and the frequency of suggestions provided by the LLM for each scenario.

\begin{figure}[tpb]
\centerline{\includegraphics[width=0.5\textwidth]{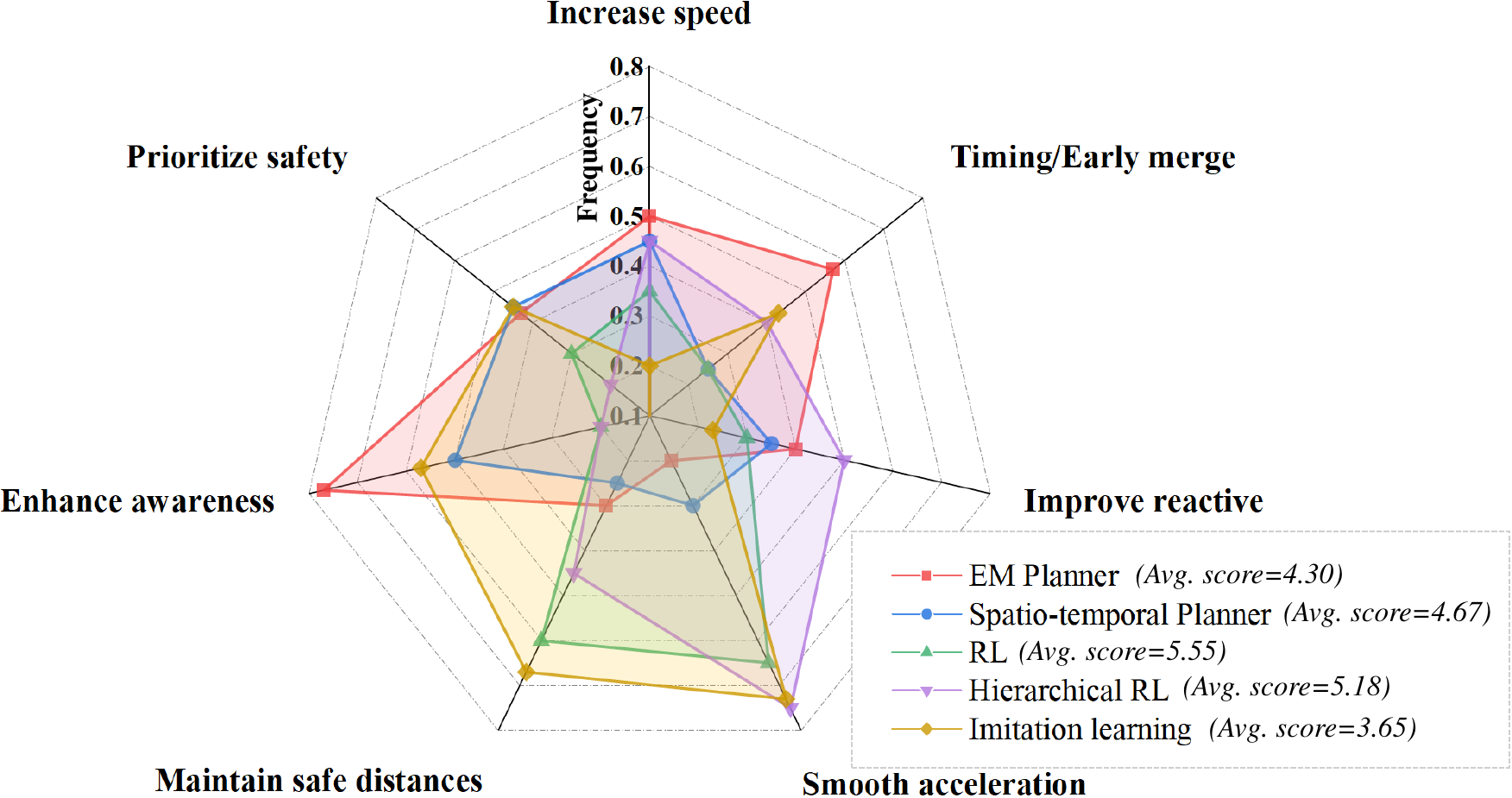}}
\caption{The performance of different methods implemented in Bench4Merge, focusing on the average scores obtained and the frequency of those suggestions were proposed.}
\label{fig:7}
\end{figure}

 As shown in Figure 7, we can draw the following conclusions: First, we observed that none of the methods performed well across all aspects. The {RL}-based method\cite{chen2023deep} achieved the highest score but exhibited issues with sharp acceleration changes. In contrast, the Imitation Learning method \cite{zhai2023rethinking} had the lowest score but demonstrated advantages in passing speed. Second, the suggestion “Enhance awareness” was the most frequently given for optimization-based methods \cite{fan2018baidu,li2022autonomous}, indicating a significant lack of interaction with surrounding vehicles. On the other hand, “Smooth Acceleration” was the most common suggestion for neural network-based methods \cite{brito2022learning,chen2023deep,zhai2023rethinking}, highlighting a deficiency in comfort. Finally, Bench4Merge also revealed issues that had not been identified in previous work, Hierarchical RL have shown significant advantages over Spatio-temporal Planner \cite{li2022autonomous} in previous work \cite{brito2022learning}. However, in Bench4Merge, the average score reflects only a marginal advantage and is lower than that of RL method. This is because prior metrics did not account for comfort, whereas Bench4Merge offers a more comprehensive evaluation. 


\section{Conclusion}
In this work, we introduced the Bench4Merge for dense merging scenarios, aimed at evaluating the performance of motion planning methods comprehensively. We extracted initial scenarios from large-scale data and classify them accordingly and trained surrounding vehicles to exhibit realistic and diverse interaction behaviors. Finally, we redefine the evaluation mechanism by leveraging an LLM to score and analyze each sample. This evaluation approach overcoming the limitations of previous evaluation methods. From the analysis, we have distilled valuable analyses from various methods, and have open-sourced all environments and algorithms. Future work will focus on generating 3D environments for dense interaction scenarios with scene rendering, catering to the closed-loop testing needs of the increasingly prominent multimodal end-to-end autonomous driving systems.

\section*{ACKNOWLEDGEMENTS}

This work is supported by Wuxi Research Institute of Applied Technologies, Tsinghua University under Grant 20242001120 and Didi Chuxing.


\bibliographystyle{IEEEtran}
\bibliography{References}

\begin{thebibliography}{10}
\providecommand{\url}[1]{#1}
\csname url@samestyle\endcsname
\providecommand{\newblock}{\relax}
\providecommand{\bibinfo}[2]{#2}
\providecommand{\BIBentrySTDinterwordspacing}{\spaceskip=0pt\relax}
\providecommand{\BIBentryALTinterwordstretchfactor}{4}
\providecommand{\BIBentryALTinterwordspacing}{\spaceskip=\fontdimen2\font plus
\BIBentryALTinterwordstretchfactor\fontdimen3\font minus \fontdimen4\font\relax}
\providecommand{\BIBforeignlanguage}[2]{{%
\expandafter\ifx\csname l@#1\endcsname\relax
\typeout{** WARNING: IEEEtran.bst: No hyphenation pattern has been}%
\typeout{** loaded for the language `#1'. Using the pattern for}%
\typeout{** the default language instead.}%
\else
\language=\csname l@#1\endcsname
\fi
#2}}
\providecommand{\BIBdecl}{\relax}
\BIBdecl

\bibitem{knaup2024active}
J.~Knaup, J.~D’sa, B.~Chalaki, T.~Naes, H.~N. Mahjoub, E.~Moradi-Pari, and P.~Tsiotras, ``Active learning with dual model predictive path-integral control for interaction-aware autonomous highway on-ramp merging,'' in \emph{2024 IEEE International Conference on Robotics and Automation (ICRA)}.\hskip 1em plus 0.5em minus 0.4em\relax IEEE, 2024, pp. 14\,191--14\,197.

\bibitem{hou2024merging}
X.~Hou, M.~Gan, W.~Wu, C.~Wang, Y.~Ji, and S.~Zhao, ``Merging planning in dense traffic scenarios using interactive safe reinforcement learning,'' \emph{Knowledge-Based Systems}, vol. 290, p. 111548, 2024.

\bibitem{brito2022learning}
B.~Brito, A.~Agarwal, and J.~Alonso-Mora, ``Learning interaction-aware guidance for trajectory optimization in dense traffic scenarios,'' \emph{IEEE Transactions on Intelligent Transportation Systems}, vol.~23, no.~10, pp. 18\,808--18\,821, 2022.

\bibitem{bouton2019cooperation}
M.~Bouton, A.~Nakhaei, K.~Fujimura, and M.~J. Kochenderfer, ``Cooperation-aware reinforcement learning for merging in dense traffic,'' in \emph{2019 IEEE Intelligent Transportation Systems Conference (ITSC)}.\hskip 1em plus 0.5em minus 0.4em\relax IEEE, 2019, pp. 3441--3447.

\bibitem{ni2016vehicle}
D.~Ni, J.~D. Leonard, C.~Jia, and J.~Wang, ``Vehicle longitudinal control and traffic stream modeling,'' \emph{Transportation Science}, vol.~50, no.~3, pp. 1016--1031, 2016.

\bibitem{xu2023bits}
D.~Xu, Y.~Chen, B.~Ivanovic, and M.~Pavone, ``Bits: Bi-level imitation for traffic simulation,'' in \emph{2023 IEEE International Conference on Robotics and Automation (ICRA)}.\hskip 1em plus 0.5em minus 0.4em\relax IEEE, 2023, pp. 2929--2936.

\bibitem{gulino2023waymax}
C.~Gulino, J.~Fu, W.~Luo, G.~Tucker, E.~Bronstein, Y.~Lu, J.~Harb, X.~Pan, Y.~Wang, X.~Chen \emph{et~al.}, ``Waymax: An accelerated, data-driven simulator for large-scale autonomous driving research,'' \emph{Advances in Neural Information Processing Systems}, vol.~36, pp. 7730--7742, 2023.

\bibitem{suo2021trafficsim}
S.~Suo, S.~Regalado, S.~Casas, and R.~Urtasun, ``Trafficsim: Learning to simulate realistic multi-agent behaviors,'' in \emph{Proceedings of the IEEE/CVF Conference on Computer Vision and Pattern Recognition}, 2021, pp. 10\,400--10\,409.

\bibitem{nishi2019merging}
T.~Nishi, P.~Doshi, and D.~Prokhorov, ``Merging in congested freeway traffic using multipolicy decision making and passive actor-critic learning,'' \emph{IEEE Transactions on Intelligent Vehicles}, vol.~4, no.~2, pp. 287--297, 2019.

\bibitem{dauner2023parting}
D.~Dauner, M.~Hallgarten, A.~Geiger, and K.~Chitta, ``Parting with misconceptions about learning-based vehicle motion planning,'' in \emph{Conference on Robot Learning}.\hskip 1em plus 0.5em minus 0.4em\relax PMLR, 2023, pp. 1268--1281.

\bibitem{hasuo2022goal}
I.~Hasuo, C.~Eberhart, J.~Haydon, J.~Dubut, R.~Bohrer, T.~Kobayashi, S.~Pruekprasert, X.-Y. Zhang, E.~A. Pallas, A.~Yamada \emph{et~al.}, ``Goal-aware rss for complex scenarios via program logic,'' \emph{IEEE Transactions on Intelligent Vehicles}, vol.~8, no.~4, pp. 3040--3072, 2022.

\bibitem{chandra2023meteor}
R.~Chandra, X.~Wang, M.~Mahajan, R.~Kala, R.~Palugulla, C.~Naidu, A.~Jain, and D.~Manocha, ``Meteor: A dense, heterogeneous, and unstructured traffic dataset with rare behaviors,'' in \emph{2023 IEEE International Conference on Robotics and Automation (ICRA)}.\hskip 1em plus 0.5em minus 0.4em\relax IEEE, 2023, pp. 9169--9175.

\bibitem{wang2023multiverse}
Y.~Wang, T.~Zhao, and F.~Yi, ``Multiverse transformer: 1st place solution for waymo open sim agents challenge 2023,'' \emph{arXiv preprint arXiv:2306.11868}, 2023.

\bibitem{kuutti2021weakly}
S.~Kuutti, R.~Bowden, and S.~Fallah, ``Weakly supervised reinforcement learning for autonomous highway driving via virtual safety cages,'' \emph{Sensors}, vol.~21, no.~6, p. 2032, 2021.

\bibitem{liu2022autonomous}
Q.~Liu, F.~Dang, X.~Wang, and X.~Ren, ``Autonomous highway merging in mixed traffic using reinforcement learning and motion predictive safety controller,'' in \emph{2022 IEEE 25th International Conference on Intelligent Transportation Systems (ITSC)}.\hskip 1em plus 0.5em minus 0.4em\relax IEEE, 2022, pp. 1063--1069.

\bibitem{liu2022zju}
L.~Xiao, ``Research on behavior decision, planning, and control of autonomous driving vehicles in high speed environments,'' Ph.D. dissertation, ZheJiang University, 2022.

\bibitem{burger2022interaction}
C.~Burger, J.~Fischer, F.~Bieder, {\"O}.~{\c{S}}. Ta{\c{s}}, and C.~Stiller, ``Interaction-aware game-theoretic motion planning for automated vehicles using bi-level optimization,'' in \emph{2022 IEEE 25th International Conference on Intelligent Transportation Systems (ITSC)}.\hskip 1em plus 0.5em minus 0.4em\relax IEEE, 2022, pp. 3978--3985.

\bibitem{zhang2023trafficbots}
Z.~Zhang, A.~Liniger, D.~Dai, F.~Yu, and L.~Van~Gool, ``Trafficbots: Towards world models for autonomous driving simulation and motion prediction,'' in \emph{2023 IEEE International Conference on Robotics and Automation (ICRA)}.\hskip 1em plus 0.5em minus 0.4em\relax IEEE, 2023, pp. 1522--1529.

\bibitem{zhong2023guided}
Z.~Zhong, D.~Rempe, D.~Xu, Y.~Chen, S.~Veer, T.~Che, B.~Ray, and M.~Pavone, ``Guided conditional diffusion for controllable traffic simulation,'' in \emph{2023 IEEE International Conference on Robotics and Automation (ICRA)}.\hskip 1em plus 0.5em minus 0.4em\relax IEEE, 2023, pp. 3560--3566.

\bibitem{jia2024bench2drive}
X.~Jia, Z.~Yang, Q.~Li, Z.~Zhang, and J.~Yan, ``Bench2drive: Towards multi-ability benchmarking of closed-loop end-to-end autonomous driving,'' \emph{arXiv preprint arXiv:2406.03877}, 2024.

\bibitem{zhou2024behaviorgpt}
Z.~Zhou, H.~Hu, X.~Chen, J.~Wang, N.~Guan, K.~Wu, Y.-H. Li, Y.-K. Huang, and C.~J. Xue, ``Behaviorgpt: Smart agent simulation for autonomous driving with next-patch prediction,'' \emph{arXiv preprint arXiv:2405.17372}, 2024.

\bibitem{bae2022lane}
S.~Bae, D.~Isele, A.~Nakhaei, P.~Xu, A.~M. Anon, C.~Choi, K.~Fujimura, and S.~Moura, ``Lane-change in dense traffic with model predictive control and neural networks,'' \emph{IEEE Transactions on Control Systems Technology}, vol.~31, no.~2, pp. 646--659, 2022.

\bibitem{lee2023robust}
K.~Lee, J.~Li, D.~Isele, J.~Park, K.~Fujimura, and M.~J. Kochendorfer, ``Robust driving policy learning with guided meta reinforcement learning,'' in \emph{2023 IEEE 26th International Conference on Intelligent Transportation Systems (ITSC)}.\hskip 1em plus 0.5em minus 0.4em\relax IEEE, 2023, pp. 4114--4120.

\bibitem{ding2021epsilon}
W.~Ding, L.~Zhang, J.~Chen, and S.~Shen, ``Epsilon: An efficient planning system for automated vehicles in highly interactive environments,'' \emph{IEEE Transactions on Robotics}, vol.~38, no.~2, pp. 1118--1138, 2021.

\bibitem{li2023marc}
T.~Li, L.~Zhang, S.~Liu, and S.~Shen, ``Marc: Multipolicy and risk-aware contingency planning for autonomous driving,'' \emph{IEEE Robotics and Automation Letters}, 2023.

\bibitem{gu2023exploring}
R.~Gu, Y.~Li, and X.~Cen, ``Exploring the stimulative effect on following drivers in a consecutive lane change using microscopic vehicle trajectory data,'' \emph{Transportation safety and environment}, vol.~5, no.~2, p. tdac047, 2023.

\bibitem{hao2020research}
W.~Hao, Z.~Zhang, Z.~Gao, K.~Yi, L.~Liu, and J.~Wang, ``Research on mandatory lane-changing behavior in highway weaving sections,'' \emph{Journal of advanced transportation}, vol. 2020, no.~1, p. 3754062, 2020.

\bibitem{daamen2010empirical}
W.~Daamen, M.~Loot, and S.~P. Hoogendoorn, ``Empirical analysis of merging behavior at freeway on-ramp,'' \emph{Transportation Research Record}, vol. 2188, no.~1, pp. 108--118, 2010.

\bibitem{wang2023faster}
H.~Wang, W.~Hao, J.~So, X.~Xiao, Z.~Chen, and J.~Hu, ``A faster cooperative lane change controller enabled by formulating in spatial domain,'' \emph{IEEE Transactions on Intelligent Vehicles}, 2023.

\bibitem{zhang2020efficient}
L.~Zhang, W.~Ding, J.~Chen, and S.~Shen, ``Efficient uncertainty-aware decision-making for automated driving using guided branching,'' in \emph{2020 IEEE International Conference on Robotics and Automation (ICRA)}.\hskip 1em plus 0.5em minus 0.4em\relax IEEE, 2020, pp. 3291--3297.

\bibitem{dauner2024navsim}
D.~Dauner, M.~Hallgarten, T.~Li, X.~Weng, Z.~Huang, Z.~Yang, H.~Li, I.~Gilitschenski, B.~Ivanovic, M.~Pavone \emph{et~al.}, ``Navsim: Data-driven non-reactive autonomous vehicle simulation and benchmarking,'' \emph{arXiv preprint arXiv:2406.15349}, 2024.

\bibitem{zhang2023ad4che}
Y.~Zhang, C.~Wang, R.~Yu, L.~Wang, W.~Quan, Y.~Gao, and P.~Li, ``The ad4che dataset and its application in typical congestion scenarios of traffic jam pilot systems,'' \emph{IEEE Transactions on Intelligent Vehicles}, vol.~8, no.~5, pp. 3312--3323, 2023.

\bibitem{lao2012gaussian}
Y.~Lao, G.~Zhang, J.~Corey, and Y.~Wang, ``Gaussian mixture model-based speed estimation and vehicle classification using single-loop measurements,'' \emph{Journal of intelligent transportation systems}, vol.~16, no.~4, pp. 184--196, 2012.

\bibitem{krajewski2018highd}
R.~Krajewski, J.~Bock, L.~Kloeker, and L.~Eckstein, ``The highd dataset: A drone dataset of naturalistic vehicle trajectories on german highways for validation of highly automated driving systems,'' in \emph{2018 21st international conference on intelligent transportation systems (ITSC)}.\hskip 1em plus 0.5em minus 0.4em\relax IEEE, 2018, pp. 2118--2125.

\bibitem{caesar2021nuplan}
H.~Caesar, J.~Kabzan, K.~S. Tan, W.~K. Fong, E.~Wolff, A.~Lang, L.~Fletcher, O.~Beijbom, and S.~Omari, ``nuplan: A closed-loop ml-based planning benchmark for autonomous vehicles,'' \emph{arXiv preprint arXiv:2106.11810}, 2021.

\bibitem{sha2023languagempc}
H.~Sha, Y.~Mu, Y.~Jiang, L.~Chen, C.~Xu, P.~Luo, S.~E. Li, M.~Tomizuka, W.~Zhan, and M.~Ding, ``Languagempc: Large language models as decision makers for autonomous driving,'' \emph{arXiv preprint arXiv:2310.03026}, 2023.

\bibitem{Deepseekr1}
\BIBentryALTinterwordspacing
(2025) deepseek. [Online]. Available: \url{https://www.deepseek.com/}
\BIBentrySTDinterwordspacing

\bibitem{chen2024driving}
L.~Chen, O.~Sinavski, J.~H{\"u}nermann, A.~Karnsund, A.~J. Willmott, D.~Birch, D.~Maund, and J.~Shotton, ``Driving with llms: Fusing object-level vector modality for explainable autonomous driving,'' in \emph{2024 IEEE International Conference on Robotics and Automation (ICRA)}.\hskip 1em plus 0.5em minus 0.4em\relax IEEE, 2024, pp. 14\,093--14\,100.

\bibitem{asuero2006correlation}
A.~G. Asuero, A.~Sayago, and A.~Gonz{\'a}lez, ``The correlation coefficient: An overview,'' \emph{Critical reviews in analytical chemistry}, vol.~36, no.~1, pp. 41--59, 2006.

\bibitem{chen2023deep}
D.~Chen, M.~R. Hajidavalloo, Z.~Li, K.~Chen, Y.~Wang, L.~Jiang, and Y.~Wang, ``Deep multi-agent reinforcement learning for highway on-ramp merging in mixed traffic,'' \emph{IEEE Transactions on Intelligent Transportation Systems}, vol.~24, no.~11, pp. 11\,623--11\,638, 2023.

\bibitem{zhai2023rethinking}
J.-T. Zhai, Z.~Feng, J.~Du, Y.~Mao, J.-J. Liu, Z.~Tan, Y.~Zhang, X.~Ye, and J.~Wang, ``Rethinking the open-loop evaluation of end-to-end autonomous driving in nuscenes,'' \emph{arXiv preprint arXiv:2305.10430}, 2023.

\bibitem{fan2018baidu}
H.~Fan, F.~Zhu, C.~Liu, L.~Zhang, L.~Zhuang, D.~Li, W.~Zhu, J.~Hu, H.~Li, and Q.~Kong, ``Baidu apollo em motion planner,'' \emph{arXiv preprint arXiv:1807.08048}, 2018.

\bibitem{li2022autonomous}
B.~Li, Y.~Ouyang, L.~Li, and Y.~Zhang, ``Autonomous driving on curvy roads without reliance on frenet frame: A cartesian-based trajectory planning method,'' \emph{IEEE Transactions on Intelligent Transportation Systems}, vol.~23, no.~9, pp. 15\,729--15\,741, 2022.

\end{thebibliography}


\appendix

\section*{Appendix A: Training Data Construction}

\begin{figure}
\centerline{\includegraphics[width=0.5\textwidth]{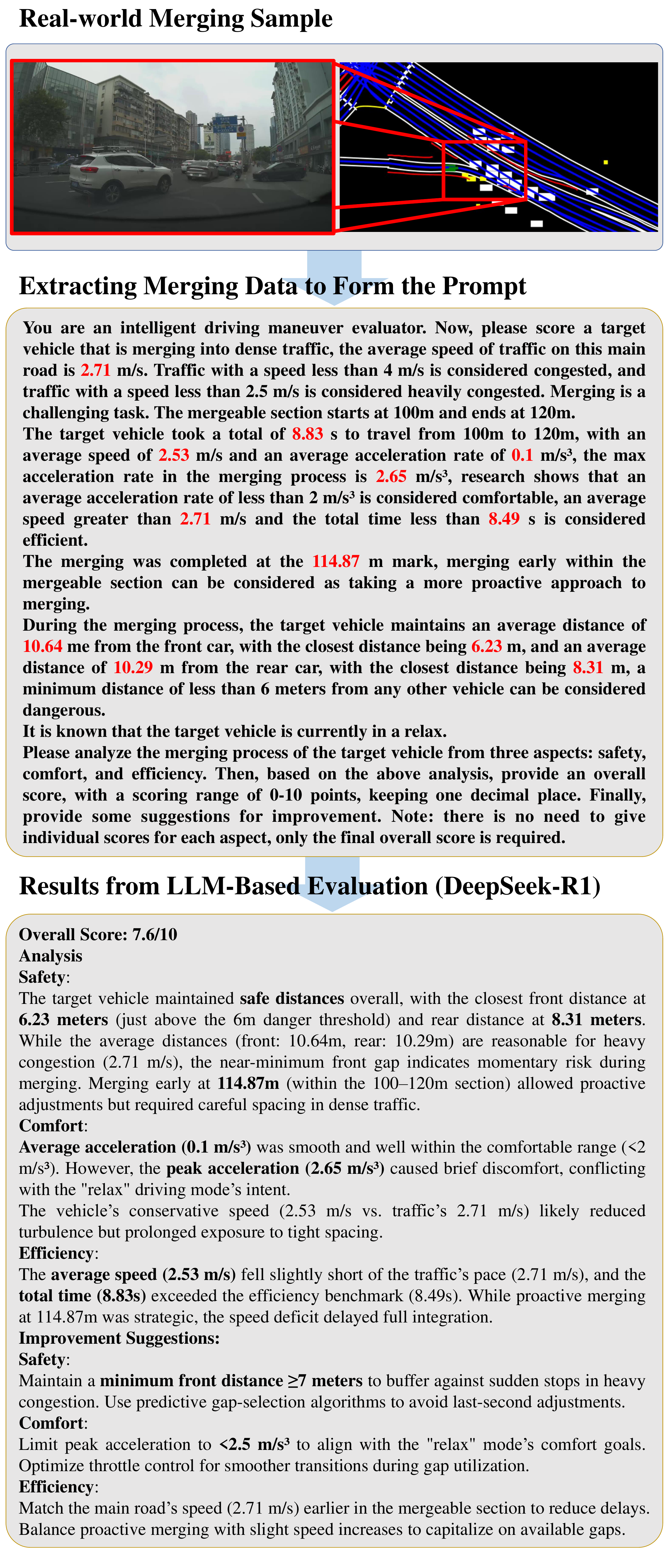}}
\caption{The evaluation of real-world autonomous driving platform based on the LLM provides valuable guidance for the planning }
\label{fig:8}
\end{figure}

Based on the definitions used in existing work for dense traffic scenes, we set the following requirements to filter the data: the Target Vehicle and the Leading Vehicle must be on the same lane, with a distance between them no greater than 10 meters, and their speeds must range from 1 $m/s$ to 5 $m/s$. Additionally, there must be interacting vehicles within a 5-meter range both in front and behind the Target Vehicle in the adjacent lanes.

Finally, our data composition is as follows: The DJI dataset contains 270,000 samples, while the nuPlan and ExiD datasets each contain 100,000 samples, we also used 100,000 samples collected from the DiDi test vehicle. Each sample is composed of four components: the Target vehicle, the Leading Vehicle, the Interactive Vehicle, and the map information, as shown in the Fig.~\ref{fig:5}. The data for Target Vehicle is represented as an 11-dimensional vector, The dataset includes the following vehicle parameters: lateral position, longitudinal position, heading angle, lateral speed, longitudinal speed, lateral acceleration, longitudinal acceleration, total acceleration, steering angle, distance to the preceding vehicle (Thw), offset distance from the lane centerline (Offset), label (as introduced in Section\uppercase\expandafter{\romannumeral3}-C), and length.

\section*{Appendix B:  Example of LLM-Based Evaluation}

We have deployed the evaluation mechanism onto our test vehicle. Fig.~\ref{fig:8} illustrates one evaluation process of the system. The input to the LLM consists of four main components: First, all the data related to the ego vehicle's merging process, including average speed, total time, the coordinates of the merging point, acceleration fluctuations, and the average and minimum distances maintained with adjacent vehicles. Second, the main lane traffic data, which includes the average speed of vehicles on the main road. Third, prior knowledge set based on experience, which includes the definition of comfortable acceleration and the range of efficient passing times. Fourth, the status of the ego vehicle, which can be selected from either relax or hurry.

\end{document}